\documentclass[10pt,twocolumn,letterpaper]{article}

\usepackage{cvpr}              % To produce the CAMERA-READY version
\definecolor{cvprblue}{rgb}{0.21,0.49,0.74}
\usepackage[pagebackref,breaklinks,colorlinks,allcolors=cvprblue]{hyperref}
\usepackage{color}
\usepackage[most]{tcolorbox}
\usepackage[utf8]{inputenc}
\usepackage[T1]{fontenc}
\usepackage{booktabs}
\usepackage{array}
\usepackage{multirow}
\usepackage{xcolor}
\usepackage{colortbl}
\usepackage{graphicx}

\definecolor{lightgray}{gray}{0.92}
\title{AdaZoom-GUI: Adaptive Zoom-based GUI Grounding with Instruction Refinement}

\author{Siqi Pei\\
Lenovo Research\\
Beijing, China\\
{\tt\small peisq2@lenovo.com}
\and
Liang Tang\\
Lenovo Research\\
Beijing, China\\
{\tt\small tangliang5@lenovo.com}
\and
Tiaonan Duan\\
Lenovo Research\\
Beijing, China\\
{\tt\small duantn1@lenovo.com}
\and
Long Chen\\
Lenovo Research\\
Beijing, China\\
{\tt\small chenlong5@lenovo.com}
\and
Shuxian Li\\
Lenovo Research\\
Beijing, China\\
{\tt\small lisx14@lenovo.com}
\and
Kaer Huang\\
Lenovo Research\\
Beijing, China\\
{\tt\small huangke1@lenovo.com}
\and
Yanzhe Jing\\
Lenovo Research\\
Beijing, China\\
{\tt\small jingyz1@lenovo.com}
\and
Yiqiang Yan\\
Lenovo Research\\
Beijing, China\\
{\tt\small yanyq@lenovo.com}
\and
Bo Zhang\\
Tsinghua University\\
Beijing, China\\
{\tt\small bo-zhang23@mails.tsinghua.edu.cn}
\and
Chenghao Jiang\\
Tsinghua University\\
Beijing, China\\
{\tt\small chenghao.jiang2022@outlook.com}
\and
Borui Zhang\\
Tsinghua University\\
Beijing, China\\
{\tt\small zhang-br21@mails.tsinghua.edu.cn}
\and
Jiwen Lu\\
Tsinghua University\\
Beijing, China\\
{\tt\small lujiwen@tsinghua.edu.cn}
}

\begin{document}
\maketitle
\begin{abstract}

GUI grounding is a critical capability for vision-language models (VLMs) that enables automated interaction with graphical user interfaces by locating target elements from natural language instructions. However, grounding on GUI screenshots remains challenging due to high-resolution images, small UI elements, and ambiguous user instructions. In this work, we propose AdaZoom-GUI, an adaptive zoom-based GUI grounding framework that improves both localization accuracy and instruction understanding. Our approach introduces an instruction refinement module that rewrites natural language commands into explicit and detailed descriptions, allowing the grounding model to focus on precise element localization. In addition, we design a conditional zoom-in strategy that selectively performs a second-stage inference on predicted small elements, improving localization accuracy while avoiding unnecessary computation and context loss on simpler cases. To support this framework, we construct a high-quality GUI grounding dataset and train the grounding model using Group Relative Policy Optimization (GRPO), enabling the model to predict both click coordinates and element bounding boxes. Experiments on public benchmarks demonstrate that our method achieves state-of-the-art performance among models with comparable or even larger parameter sizes, highlighting its effectiveness for high-resolution GUI understanding and practical GUI agent deployment.

\end{abstract}    
\section{Introduction}

\paragraph{}
% Graphical User Interfaces (GUIs) are designed as a medium to enhance user experience during human-device interaction, which neglects the need for automated processes. In a wide range of cases, GUI is the only way to perceive information and interact with a device, making language model based agents interacting with devices by interpreting structured texts, e.g., HTML, a11y tree, fail in this case \cite{kim2023language}. To solve this issue, extensive research has been done on GUI agents, trying to automate the interaction with GUIs with given instruction \cite{wang2024gui}. GUI grounding is an ability for vision-language models to understand human's natural language instructions and locate the target elements in screenshots. As a core ability for GUI agent, it potentially changes the way of human interaction with GUI-based devices. By converting vague instructions to precise coordinates, GUI grounding can automate those operations only accessible through GUI interaction, which suffers from unfixed element location, variable element shape, etc.

Graphical User Interfaces (GUIs) are designed to enhance user experience in human-device interaction, but they often overlook the need for automated processes. In many scenarios, GUIs provide the only interface through which users can perceive information and interact with devices. This makes it difficult for language-model-based agents that rely on structured representations such as HTML or accessibility trees (a11y tree) to interact with such systems effectively \cite{kim2023language}. To address this limitation, extensive research has explored GUI agents that aim to automate interactions with graphical interfaces based on natural language instructions \cite{wang2024gui}. 

GUI grounding refers to the capability of vision-language models (VLMs) to understand natural language instructions and locate the corresponding target elements within GUI screenshots. As a fundamental capability for GUI agents, GUI grounding has the potential to transform the way humans interact with GUI-based devices. By converting vague instructions into precise coordinates, it enables automated execution of tasks that are otherwise accessible only through GUI interactions, where element locations and shapes may vary across interfaces.

% As the development of Vision-language models (VLMs), general VLMs \cite{gemini3pro_modelcard,Qwen3-VL} are already capable for some GUI grounding tasks, and most existing works \cite{cheng2024seeclick,tang2025gui,yang2025gta1} are built upon VLMs and trained with supervised fine-tuning (SFT) and/or reinforcement learning. Although these methods perform well in certain scenarios, they still face challenge in complex samples, especially with higher-resolution GUIs or with instructions that requires stronger comprehension ability.

With the rapid development of vision-language models, general-purpose VLMs \cite{gemini3pro_modelcard,Qwen3-VL} have demonstrated promising abilities on GUI grounding tasks. Many existing approaches \cite{cheng2024seeclick,tang2025gui,yang2025gta1} build upon these models and further train them using supervised fine-tuning (SFT) and/or reinforcement learning. Although these methods achieve strong performance in certain scenarios, they still struggle with complex cases, particularly when dealing with high-resolution GUIs or instructions that require stronger semantic understanding.

% Compared to images from other domains, GUI screenshots have a much higher resolution, and the much smaller relative icon and text size introduces high information loss during downsample, making grounding models hard to accurately locate small elements in a screenshot. To address this problem, some works \cite{park2025r,zhou2025mai} designed a manual two-step strategy, which zooms-in the screenshot around the prediction point with a fixed ratio. However, this fixed two-stage strategy brings additional unnecessary computation to simple or low resolution screenshots, which may even reduce the performance due to the loss of context information outside the zoomed-in region.

Compared with images from other domains, GUI screenshots typically have much higher resolutions, while UI elements such as icons and texts occupy relatively small regions. This leads to significant information loss during downsampling, making it difficult for grounding models to accurately localize small elements. To address this issue, some prior works \cite{park2025r,zhou2025mai} adopt a two-stage strategy that zooms in on the screenshot around a predicted point using a fixed ratio. However, this fixed two-stage design introduces unnecessary computation for simple or low-resolution screenshots, and may even degrade performance due to the loss of contextual information outside the zoomed-in region.

% To alleviate this issue, we proposed a conditioned zoomed-in strategy which selectively apply the zoom-in method only to necessary GUI screenshots. Namely, we constructed a high-quality dataset and trained a model using Group Relative Policy Optimization (GRPO) \cite{shao2024deepseekmath} to predict both the location and size of target elements - the only two information normally contained in training datasets. Only those scenarios with relatively small target element are then applied with the zoom-in strategy. In this way, our method can intelligently apply fast single-stage inference to simple cases, and accurate two-stage inference to complex scenarios.

To alleviate this problem, we propose a \textit{conditional zoom-in strategy} that selectively applies zoom-in operations only when necessary. Specifically, we construct a high-quality GUI grounding dataset and train a model using Group Relative Policy Optimization (GRPO) \cite{shao2024deepseekmath} to predict both the location and size of target elements — the two pieces of information typically provided in grounding annotations. The zoom-in strategy is triggered only when the predicted target element is relatively small. In this way, the model can perform efficient single-stage inference for simple cases while applying more accurate two-stage inference for complex scenarios.

% For most training data \cite{kapoor2024omniact,yang2025gta1,gou2024navigating}, the natural language instruction is always simple and straightforward, hight quality instruction data is rare due to human labor cost. However, in real world use case, human instruction might be complex and vague, requiring model's comprehension ability. We argue that a model's comprehension ability and grounding ability are two independent and distinct abilities, so training on grounding dataset with simple instructions can only improve the grounding ability and might degrade the comprehension ability. To solve this problem, we applied another model before the grounding model, to refine the instruction into a clear one with detailed description of the target element. In this way, the grounding model can be focused on answering "which exact point to click", but not comprehending "which element to click".

Another limitation of existing datasets \cite{kapoor2024omniact,yang2025gta1,gou2024navigating} is that their natural language instructions are usually simple and straightforward, as high-quality instruction annotations are expensive to obtain. In real-world applications, however, user instructions are often complex or ambiguous, requiring stronger semantic understanding. We argue that instruction comprehension and GUI grounding are two distinct capabilities. Training a model solely on grounding datasets with simple instructions may improve localization ability but does not necessarily enhance instruction understanding. 

To address this issue, we introduce an instruction refinement model that rewrites the original instruction into a clearer and more detailed description of the target element before passing it to the grounding model. This design allows the grounding model to focus on answering \textit{“which exact point to click”} rather than interpreting \textit{“which element should be clicked”}.

The contributions of this work are summarized as follows:
\begin{itemize}
    % \item We propose \textcolor{red}{xxx}, a novel GUI grounding frameworks that adapts conditional zoom-in strategy, with a instruction refining model to let the grounding model focus on grounding but not comprehension. It can intelligently apply fast single-stage inference to simple cases, and accurate two-stage inference to complex scenarios.

    \item We propose AdaZoom-GUI, a novel GUI grounding framework that integrates an instruction refinement module with a conditional zoom-in strategy. This design enables the model to apply efficient single-stage inference for simple cases and accurate two-stage inference for complex scenarios.

    % \item We construct a high quality GUI grounding dataset and trained the grounding model using GRPO, which can predict both the location and size of target elements, enabling the conditional zoom-in strategy.

    \item We construct a high-quality GUI grounding dataset and train the grounding model using GRPO, enabling the model to predict both the location and size of target elements, which further supports the conditional zoom-in strategy.

    % \item We evaluate the model on public benchmarks, showing that our model achieves state-of-the-art performance among models with similar or even larger size.

    \item We evaluate our method on public benchmarks and demonstrate that it achieves state-of-the-art performance among models with comparable or even larger parameter sizes.
\end{itemize}
\section{Method}

\begin{figure*}[t]
  \centering
  \includegraphics[width=\linewidth]{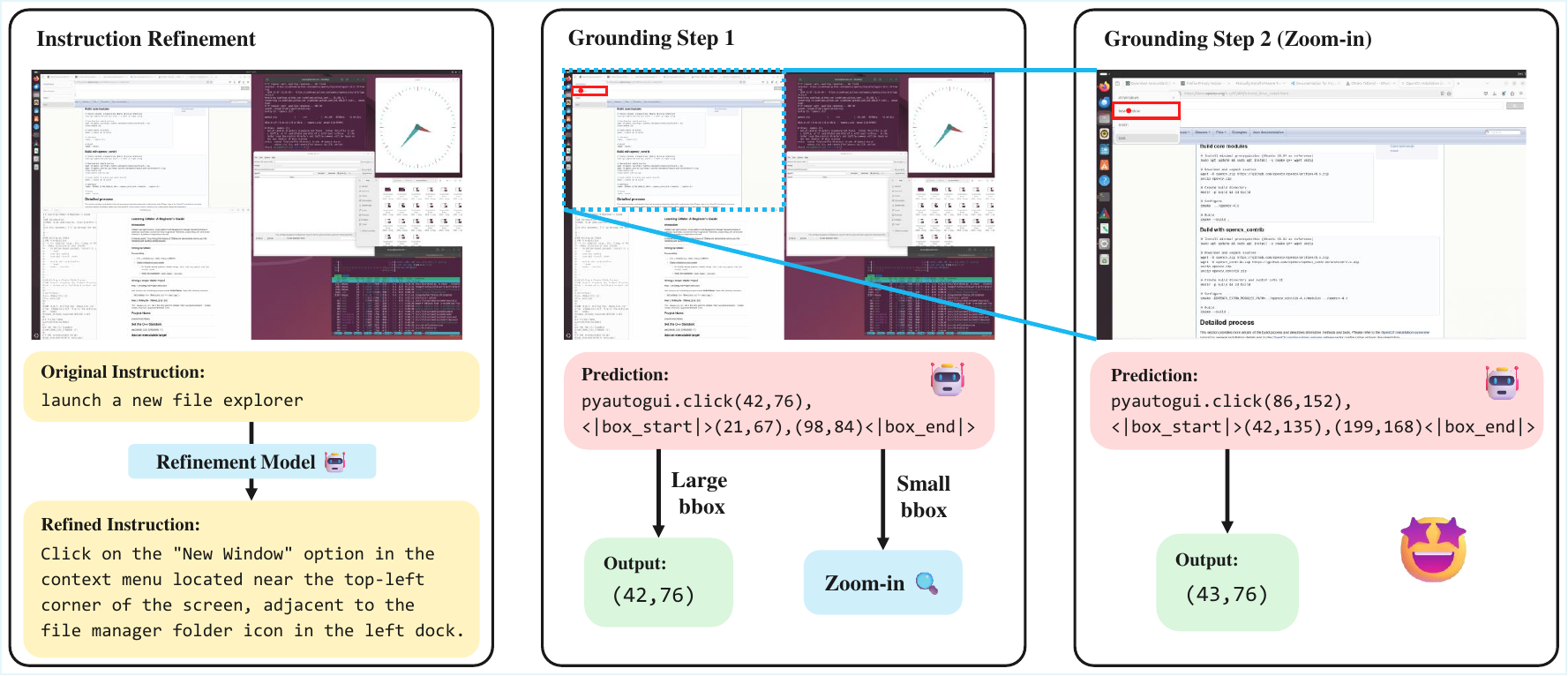}
  \caption{Overall grounding inference pipeline. Given a GUI screenshot and a natural language instruction, a refinement model first rewrites the instruction into a direct and detailed command. The grounding model then takes the rewritten instruction and the screenshot as input and outputs both the click-point and the bounding box of the target element. If the predicted bounding box is larger than a certain threshold, the click-point is directly used as the final output. Otherwise, the screenshot is zoomed in around the click-point and fed into the grounding model again for a second round of inference.
  }
  \label{fig:overall_pipeline}
\end{figure*}

\paragraph{}
The overall grounding pipeline of AdaZoom-GUI is illustrated in Figure~\ref{fig:overall_pipeline}. It consists of two components: an instruction refinement model and a grounding model. The details of each component are described in the following sections.

\subsection{Instruction Refinement}

\paragraph{}
% The instruction refinement model is designed to rewrite the original natural language instruction into a more direct and detailed command that can be easily understood by the grounding model. This is particularly important for complex instructions that may contain indirect references or ambiguous language. The refinement model takes the original instruction and the GUI screenshot as input and generates a refined instruction that explicitly describes the target element and the desired action. The model should refines the instruction on two dimensions: 

The instruction refinement model is designed to rewrite the original natural language instruction into a more explicit and detailed command that can be more easily understood by the grounding model. This is particularly important for complex instructions that may contain indirect references or ambiguous expressions. 

The refinement model takes the original instruction and the GUI screenshot as input and generates a refined instruction that explicitly describes the target element and the desired action. Specifically, the model refines the instruction along two dimensions:

\begin{enumerate}
    % \item Interpret the instruction and locate the exact target element to be grounded. For example, if the instruction is "launch a new file explorer", the refinement model will interpret it as "click on the "New Window" option", which convert the high-level instruction into a specific element to be recognized on the GUI.
    \item \textbf{Instruction interpretation.} The model first interprets the instruction and identifies the exact target element to be grounded. For example, if the instruction is "launch a new file explorer", the refinement model may interpret it as "click on the 'New Window' option", converting a high-level instruction into a concrete GUI element.

    % \item Add more details to the instruction to make it more explicit. For example, if the instruction is "click on the download icon", the refinement model will interpret it as "click on the download icon (a downward-pointing arrow) located in the top-right corner of the Firefox browser toolbar, just to the right of the address bar", which provides more details, such as the appearance and location of the target element, to help the grounding model accurately identify it.
    \item \textbf{Instruction enrichment.} The model further enriches the instruction with additional visual or spatial details. For example, if the instruction is "click on the download icon", the refinement model may rewrite it as "click on the download icon (a downward-pointing arrow) located in the top-right corner of the Firefox browser toolbar, just to the right of the address bar". These additional descriptions, such as visual appearance or approximate location, help the grounding model identify the target element more accurately.
\end{enumerate}

To achieve this, we employ a pre-trained vision-language model with the following prompt template:

\begin{tcolorbox}[colback=gray!10, colframe=gray!50, title=Refinement prompt, breakable]
    \textbf{System prompt:}\\
    You are a helpful GUI assistant.\\
    \\
    \textbf{User prompt:}\\
    You are given a task description and a screenshot of a GUI. The task can be completed with only one click.\\
    You need to find out the target to click, and then refine the task description to let user easily locate the target on the screen.\\
    Possible refinements include adding location information, describing visual features (color, size, text, icon shape, ...), clarifying ambiguous terms, etc.\\
    \\
    Only reply with the refined description. Do not add explanations.\\
    \\
    Task: \textit{\{instruction\}}
\end{tcolorbox}

\subsection{Conditional Zoom-in Grounding}

\paragraph{}
% The grounding model is responsible for identifying the target element on the GUI based on the refined instruction. It takes the refined instruction and the GUI screenshot as input and outputs both the click-point and bounding box of the target element. The click-point is a specific coordinate on the GUI where the user should click, while the bounding box is a rectangular area that encompasses the target element. The prompt template for the grounding model is as follows:

The grounding model is responsible for locating the target element on the GUI based on the refined instruction. It takes the refined instruction and the GUI screenshot as input and outputs both the click-point and the bounding box of the target element. The click-point represents the exact coordinate where the user should click, while the bounding box indicates the spatial extent of the target element.

The prompt template for the grounding model is as follows:

\begin{tcolorbox}[colback=gray!10, colframe=gray!50, title=Grounding prompt, breakable]
    \textbf{System prompt:}\\
    You are a GUI agent. You are given a task and a screenshot of the screen. You need to perform pyautogui click/moveTo action to complete the task, and then provide the bouding box of the target object. The answer format is \texttt{\`}pyautogui.click(x=?, y=?), \textless{}\textbar{}box\_start\textbar{}\textgreater{}(x1,y1),(x2,y2)\textless{}\textbar{}box\_end\textbar{}\textgreater{}\texttt{\`}. If the task is infeasible (e.g., the task is already completed, the target does not exist in the image, or the instruction is unrelated to the screenshot), output a null action exactly as follows: \texttt{\`}pyautogui.click(x=0, y=0), \textless{}\textbar{}box\_start\textbar{}\textgreater{}(0,0),(0,0)\textless{}\textbar{}box\_end\textbar{}\textgreater{}\texttt{\`}.\\
    \\
    \textbf{User prompt:}\\
    Please complete the following tasks by clicking using \texttt{\`}pyautogui.click\texttt{\`} and returning the bounding box:\\
    Task: \textit{\{refined\_instruction\}}
\end{tcolorbox}

\subsubsection{Dataset Construction}

\paragraph{}
% To train the grounding model, we construct a dataset of GUI screenshots paired with natural language instructions and their corresponding bounding boxes. We collect a diverse set of GUI screenshots from various applications, and annotate them with natural language instructions that describe actions to be performed on the GUI. Each instruction is associated with a bounding box that indicates the location of the target element on the screenshot. We then use LLM to augment the instruction, making it with or without the location information, intention information, etc., to increase the diversity of the instruction and improve the robustness of the grounding model. The screenshots are also augmented with padding and resizing to ensure that the model can handle different screen resolutions.

To train the grounding model, we construct a dataset consisting of GUI screenshots paired with natural language instructions and their corresponding bounding boxes. We collect a diverse set of screenshots from various applications and annotate them with instructions describing the intended actions. Each instruction is associated with a bounding box indicating the location of the target element.

To further increase instruction diversity and improve model robustness, we use a large language model (LLM) to augment the instructions by generating variations with or without location descriptions, intention cues, or other contextual information. In addition, the screenshots are augmented with padding and resizing so that the model can handle different screen resolutions.

\subsubsection{Training Details}

\paragraph{}
% We use GRPO~\cite{shao2024deepseekmath} to train the grounding model. The reward function contains two components: a point reward $\mathcal{R}_{point}$ and a bounding box reward $\mathcal{R}_{bbox}$:

We train the grounding model using Group Relative Policy Optimization (GRPO)~\cite{shao2024deepseekmath}. The reward function consists of two components: a point reward $\mathcal{R}_{point}$ and a bounding box reward $\mathcal{R}_{bbox}$:

\begin{equation}
    \mathcal{R} = \lambda \mathcal{R}_{point} + (1 - \lambda) \mathcal{R}_{bbox} \ \ (0 \leq \lambda \leq 1)
\end{equation}

% The point reward $\mathcal{R}_{point}$ includes a format reward $\mathcal{R}_{f, point}$ and a point-in-box reward $\mathcal{R}_{point-in-box}$. The format reward is uesd to to verify whether the output format of the click-point is correct, with means to check if the output follows the required format of \texttt{\`}pyautogui.click(x=?, y=?), ...\texttt{\`} and if the coordinates are valid. The point-in-box reward is used to check whether the click-point falls within the bounding box of the target element, which can be calculated as:

The point reward $\mathcal{R}_{point}$ includes a format reward $\mathcal{R}_{f, point}$ and a point-in-box reward $\mathcal{R}_{point-in-box}$. The format reward verifies whether the output format of the click-point is correct, ensuring that the prediction follows the required format \texttt{\`}pyautogui.click(x=?, y=?), ...\texttt{\`} and that the coordinates are valid.

The point-in-box reward checks whether the predicted click-point lies within the ground-truth bounding box:

\begin{equation}
    \mathcal{R}_{point-in-box} = \begin{cases}
        1, & \text{if } x_1 \leq \hat{x} \leq x_2 \text{ and } y_1 \leq \hat{y} \leq y_2 \\
        0, & \text{otherwise}
    \end{cases}
\end{equation}

% Where $(\hat{x}, \hat{y})$ is the predicted click-point, and $(x_1, y_1)$ and $(x_2, y_2)$ are the top-left and bottom-right coordinates of the ground truth bounding box, respectively.

where $(\hat{x}, \hat{y})$ denotes the predicted click-point, and $(x_1, y_1)$ and $(x_2, y_2)$ are the coordinates of the ground-truth bounding box.

% The bounding box reward $\mathcal{R}_{bbox}$ similarly includes a format reward $\mathcal{R}_{f, bbox}$ and an IoU reward $\mathcal{R}_{IoU}$. The format reward checks if the output format of the bounding box is correct as \texttt{\`}..., \textless{}\textbar{}box\_start\textbar{}\textgreater{}(x1,y1),(x2,y2)\textless{}\textbar{}box\_end\textbar{}\textgreater{}\texttt{\`}, and if the coordinates are valid. The IoU reward calculates the Intersection over Union (IoU) between the predicted bounding box and the ground truth bounding box, which can be calculated as:

The bounding box reward $\mathcal{R}_{bbox}$ similarly consists of a format reward $\mathcal{R}_{f, bbox}$ and an IoU reward $\mathcal{R}_{IoU}$. The format reward checks whether the bounding box prediction follows the required format \texttt{\`}..., \textless{}\textbar{}box\_start\textbar{}\textgreater{}(x1,y1),(x2,y2)\textless{}\textbar{}box\_end\textbar{}\textgreater{}\texttt{\`}. The IoU reward measures the overlap between the predicted bounding box and the ground-truth bounding box:

\begin{equation}
    \mathcal{R}_{IoU} = \frac{\text{Area of Intersection}}{\text{Area of Union}}    
\end{equation}

% Where the area of intersection is the area of overlap between the predicted bounding box and the ground truth bounding box, and the area of union is the total area covered by both bounding boxes.

\subsubsection{Conditional Zoom-In Strategy}

\paragraph{}
% During inference, a conditional zoom-in strategy is employed to handle cases where the target element is small and difficult to locate in the original screenshot, and avoid computation overhead and information loss caused by zooming when the target element is large enough to be accurately located in the original screenshot. The judgment of whether to zoom-in is based on the size of the predicted bounding box $(\hat{x_1}, \hat{y_1}), (\hat{x_2}, \hat{y_2})$. If the width or height of the bounding box is smaller than a certain threshold, which indicates that the target element is small and may be hard to accurately locate, and thus the other side of the bounding box is smaller than another larger threshold, which makes sure the target element and its surrounding context are included in the zoomed-in image, then we say it meets the zoom-in condition. The zoom-in condition can be formulated as:

During inference, we employ a conditional zoom-in strategy to improve localization accuracy for small target elements while avoiding unnecessary computation for large elements. The decision of whether to apply zoom-in is based on the size of the predicted bounding box $(\hat{x_1}, \hat{y_1}), (\hat{x_2}, \hat{y_2})$.

If the width or height of the predicted bounding box is smaller than a predefined threshold, and thus the other side of the bounding box is smaller than another larger threshold, the element is considered small and may be difficult to localize accurately in the original screenshot. In this case, zoom-in is applied while ensuring that sufficient surrounding context is preserved. The zoom-in condition is formulated as:

\begin{equation}
    (\hat{w} \leq \alpha \wedge \hat{h} \leq \beta) \vee (\hat{h} \leq \alpha \wedge \hat{w} \leq \beta) \ \ (\alpha < \beta)
\end{equation}

% Where $\hat{w} = \hat{x_2} - \hat{x_1}$ and $\hat{h} = \hat{y_2} - \hat{y_1}$ are the width and height of the predicted bounding box, respectively. $\alpha$ is the smaller threshold and $\beta$ is the larger threshold. If the zoom-in condition is met, we will zoom in the screenshot around the click-point with a certain ratio, and resize it to the original input size. The zoomed-in screenshot is then fed into the grounding model again for a second round of inference, which can help the model to better locate small target elements by providing a closer view of the relevant area. This strategy can significantly improve the accuracy of the grounding model, especially for tasks that involve small icons or text elements on the GUI.

where $\hat{w} = \hat{x_2} - \hat{x_1}$ and $\hat{h} = \hat{y_2} - \hat{y_1}$ denote the predicted bounding box width and height, respectively. $\alpha$ and $\beta$ are the smaller and larger thresholds.

If the condition is satisfied, the screenshot is zoomed in around the predicted click-point with a predefined ratio and resized back to the original input resolution. The zoomed-in image is then fed into the grounding model for a second round of inference. By providing a closer view of the relevant region, this strategy improves the model's ability to localize small UI elements such as icons or text, while maintaining efficiency for simpler cases.
\begin{table*}[t]
\centering
\caption{Performance Comparison on ScreenSpot-Pro benchmark.}
\label{tab:overall_results}
\resizebox{\textwidth}{!}{
\begin{tabular}{l *{12}{c} c}
\toprule
\multirow{2}{*}{\textbf{Model}} & \multicolumn{2}{c}{\textbf{Development}} & \multicolumn{2}{c}{\textbf{Creative}} & \multicolumn{2}{c}{\textbf{CAD}} & \multicolumn{2}{c}{\textbf{Scientific}} & \multicolumn{2}{c}{\textbf{Office}} & \multicolumn{2}{c}{\textbf{OS}} & \multirow{2}{*}{\textbf{Avg.}} \\
\cmidrule(lr){2-3} \cmidrule(lr){4-5} \cmidrule(lr){6-7} \cmidrule(lr){8-9} \cmidrule(lr){10-11} \cmidrule(lr){12-13}
& Text & Icon & Text & Icon & Text & Icon & Text & Icon & Text & Icon & Text & Icon & \\
\midrule
\rowcolor{lightgray}
\multicolumn{14}{l}{\textit{\textbf{General Models}}} \\
GPT-4o \cite{hurst2024gpt} & 2.0 & 0.0 & 1.3 & 0.0 & 1.0 & 0.0 & 2.1 & 0.0 & 1.1 & 0.0 & 0.0 & 0.0 & 0.8 \\
Claude 3.7 Sonnet \cite{Claude3_7_sonnet} & - & - & - & - & - & - & - & - & - & - & - & - & 27.7 \\
Operater \cite{Operater} & 50.0 & 19.3 & 51.5 & 23.1 & 16.8 & 14.1 & 58.3 & 24.5 & 60.5 & 28.3 & 34.6 & 30.3 & 36.6 \\
Gemini-3-Pro \cite{gemini3pro_modelcard} & - & - & - & - & - & - & - & - & - & - & - & - & 72.7 \\
Seed1.8 \cite{seed2025seed1} & - & - & - & - & - & - & - & - & - & - & - & - & 73.1 \\
Qwen3-VL-4B-Instruct \cite{Qwen3-VL} & - & - & - & - & - & - & - & - & - & - & - & - & 59.5 \\
Qwen3-VL-32B-Instruct \cite{Qwen3-VL} & - & - & - & - & - & - & - & - & - & - & - & - & 57.9 \\
Qwen3-VL-235B-A22B-Instruct \cite{Qwen3-VL} & - & - & - & - & - & - & - & - & - & - & - & - & 62.0 \\
Qwen3.5-397B-A17B \cite{qwen3.5} & - & - & - & - & - & - & - & - & - & - & - & - & 65.6 \\
\midrule
\rowcolor{lightgray}
\multicolumn{14}{l}{\textit{\textbf{GUI Models}}} \\
InfiGUI-R1-3B \cite{liu2025infigui} & 51.3 & 12.4 & 44.9 & 7.0 & 33.0 & 14.1 & 58.3 & 20.0 & 65.5 & 28.3 & 43.9 & 12.4 & 35.7 \\
GUI-Actor-7B \cite{wu2025gui} & 59.1 & 15.9 & 59.6 & 16.1 & 47.7 & 9.4 & 70.1 & 25.5 & 69.5 & 41.5 & 55.1 & 19.1 & 44.6 \\
GUI-G$^2$-7B \cite{tang2025gui} & 68.8 & 17.2 & 57.1 & 15.4 & 55.8 & 12.5 & 77.1 & 24.5 & 74.0 & 32.7 & 57.9 & 21.3 & 47.5 \\
OpenCUA-7B \cite{wang2025opencua} & - & - & - & - & - & - & - & - & - & - & - & - & 50.0 \\
GTA1-7B \cite{yang2025gta1} & 66.9 & 20.7 & 62.6 & 18.2 & 53.3 & 17.2 & 76.4 & 31.8 & 82.5 & 50.9 & 48.6 & 25.9 & 50.1 \\
GUI-Owl-7B \cite{ye2025mobile} & 76.6 & 31.0 & 59.6 & 27.3 & 64.5 & 21.9 & 79.1 & 37.3 & 77.4 & 39.6 & 59.8 & 33.7 & 54.9 \\
Holo2-4B \cite{Holo2} & - & - & - & - & - & - & - & - & - & - & - & - & 57.2 \\
MAI-UI-8B \cite{zhou2025mai} & 83.8 & 52.4 & 76.3 & 33.6 & 72.6 & 35.9 & 79.9 & 37.3 & 88.7 & 60.4 & 76.6 & 49.4 & 65.8 \\
MAI-UI-8B (Zoom-in) \cite{zhou2025mai} & 78.6 & 58.6 & 78.8 & 46.9 & 80.7 & 43.8 & 86.1 & 49.1 & 88.1 & 81.1 & 76.6 & 51.7 & 70.9 \\
\midrule
\rowcolor{lightgray}
\multicolumn{14}{l}{\textit{\textbf{Ours}}} \\
AdaZoom-GUI-4B & 81.2 & 40.0 & 74.2 & 23.8 & 61.4 & 21.9 & 87.5 & 39.1 & 88.7 & 20.9 & 83.2 & 37.1 & 61.6 \\
% \quad + \textit{Qwen3.5-397B-A17B \cite{qwen3.5} Refinement} & 90.3 & 49.7 & 81.8 & 37.8 & 77.2 & 35.9 & 90.3 & 42.7 & 96.6 & 66.0 & 88.8 & 52.8 & 69.7 \\
\quad + \textit{Conditional Zoom-In} & 87.7 & 47.6 & 83.3 & 38.5 & 77.7 & 32.8 & 88.9 & 49.1 & 94.4 & 66.0 & 80.4 & 53.9 & 70.6 \\
\quad + \textit{Qwen3.5-397B-A17B \cite{qwen3.5} Refinement} & 91.6 & 57.9 & 88.9 & 49.7 & 83.2 & 39.1 & 91.0 & 51.8 & 97.2 & 81.1 & 89.7 & 60.7 & 76.8 \\
\bottomrule
\end{tabular}
}
\end{table*}

\section{Experiments}

\paragraph{}
% Extensive experiments are conducted to evaluate the performance of our proposed method. We compare our approach with several state-of-the-art methods on benchmark datasets, showcasing the effectiveness of our method in enhancing the performance of GUI agents on high-resolution screenshots and complex tasks.

We conduct extensive experiments to evaluate the effectiveness of the proposed AdaZoom-GUI. Our approach is compared with several state-of-the-art methods on benchmark datasets to demonstrate its advantages in grounding accuracy, particularly for high-resolution screenshots and complex GUI tasks.

\subsection{Experimental Setup}

\paragraph{}
% We train our model based on Qwen3-VL-4B-Instruct \cite{Qwen3-VL} using GRPO implemented in the ms-swift \cite{zhao2024swiftascalablelightweightinfrastructure} framework. The vision encoder and aligner are frozen during training, while the remaining parameters are fully fine-tuned in bfloat16 precision. Each prompt generates 8 candidate responses, optimized with 4 PPO epochs and 2 mini-batches per update. The learning rate is set to $2\times10^{-7}$, with no warm-up, and the KL penalty coefficient is 0.001. The reward ratio $\lambda$ is set to 0.5. During inference, we set the zoom-in condition threshold $\alpha$ to 200 pixels and $\beta$ to 1000 pixels, and the zoom-in ratio is 2. Qwen3.5-397B-A17B \cite{qwen3.5} is used for instruction refinement.

Our grounding model is trained based on Qwen3-VL-4B-Instruct \cite{Qwen3-VL} using GRPO implemented in the ms-swift framework \cite{zhao2024swiftascalablelightweightinfrastructure}. The instruction refinement module uses Qwen3.5-397B-A17B \cite{qwen3.5}.
% During training, the vision encoder and aligner are frozen, while the remaining parameters are fully fine-tuned in bfloat16 precision. Each prompt generates 8 candidate responses. The model is optimized using 4 PPO epochs with 2 mini-batches per update. 

% The learning rate is set to $2\times10^{-7}$ without warm-up, and the KL penalty coefficient is set to 0.001. The reward balance parameter $\lambda$ is set to 0.5. 

% During inference, the zoom-in condition thresholds are set to $\alpha = 200$ pixels and $\beta = 1000$ pixels, with a zoom-in ratio of 2. 

\subsection{Main Results}

\paragraph{}
% We evaluate our method on the ScreenSpot-Pro \cite{li2025screenspotpro} benchmark, which includes high-resolution screenshots from various professional domains. The results are summarized in Table~\ref{tab:overall_results}. Our base model achieves a competitive average score of 70.6 with the conditional zoom-in mechanism, and further refinement with Qwen3.5-Plus boosts the performance to 76.8, surpassing several state-of-the-art model with similar parameter sizes, and even competitive with larger models. These results demonstrate the effectiveness of our proposed method in enhancing the performance of GUI agents on high-resolution screenshots and complex tasks.

We evaluate our AdaZoom-GUI on the ScreenSpot-Pro \cite{li2025screenspotpro} benchmark, which contains high-resolution GUI screenshots from multiple professional domains. The results are summarized in Table~\ref{tab:overall_results}.

Our base model achieves an average score of 70.6 when equipped with the conditional zoom-in strategy. When combined with the instruction refinement module based on Qwen3.5-397B-A17B, the performance further improves to 76.8. This result surpasses several state-of-the-art GUI grounding models with comparable parameter sizes and is competitive with much larger models. These results demonstrate that the proposed framework effectively improves grounding performance on complex and high-resolution GUI scenarios.

% The effectiveness of the conditional zoom-in mechanism is even more evident in the ScreenSpot-v2 \cite{wu2024atlas} benchmark, where the resolution of screenshots is much lower than that of ScreenSpot-Pro. As shown in Table~\ref{tab:overall_results_v2}, our base model achieves a high average score of 0.943, but the performance drops significantly to 0.916 when the unconditioned zoom-in mechanism is applied, which might be due to that the tasks are simple enough to be grounded with the original screenshots, and zooming in the low-resolution screenshots leads to large information loss. In contrast, the conditional zoom-in mechanism can effectively determine when to zoom in, avoid unnecessary zoom-in and information loss, and thus achieves a higher average score of 0.945. This further validates the importance of the zoom-in condition in our framework, which can effectively balance the trade-off between information loss and detail enhancement.

The benefit of the conditional zoom-in strategy becomes more evident on the ScreenSpot-v2 benchmark \cite{wu2024atlas}, where screenshots typically have much lower resolutions than those in ScreenSpot-Pro. As shown in Table~\ref{tab:overall_results_v2}, our base model achieves an average score of 0.943. However, when an unconditioned zoom-in strategy is applied, the performance drops to 0.916. This degradation likely occurs because the tasks in ScreenSpot-v2 are relatively simple and can be accurately grounded using the original screenshots. In such cases, zooming into already low-resolution images introduces additional information loss.

In contrast, the proposed conditional zoom-in mechanism selectively applies zoom-in only when necessary, avoiding unnecessary cropping and preserving global context. As a result, it achieves a higher average score of 0.945. This observation highlights the importance of adaptive zoom-in decisions, which effectively balance the trade-off between contextual information and fine-grained visual details.

\begin{table}
\centering
\caption{Performance Comparison on ScreenSpot-v2 benchmark.}
\label{tab:overall_results_v2}
\begin{tabular}{l|c}
    \toprule
    \textbf{Model} & \textbf{Avg. Score} \\
    \midrule
    Ours-4B & 0.943 \\
    \quad + \textit{Unconditioned Zoom-In} & 0.916 \\
    \quad + \textit{Conditional Zoom-In} & 0.945 \\
    \bottomrule
\end{tabular}
\end{table}

% Our model with Qwen3.5-397B-A17B refinement only also achieves an average score of 69.7, which is already much higher than the original Qwen3.5-397B-A17B model (65.6), and the performance can be further improved to 76.8 with the help of the conditional zoom-in mechanism. By adding only an additional 4B model to this 397B model, the performance can be significantly improved, which demonstrates the effectiveness of our proposed method in enhancing the performance of general large models on specific tasks, and our method can effectively leverage the knowledge from large models to enhance the performance of smaller models. This also suggests that our method can be a cost-effective way to boost the performance of existing large models without the need for extensive fine-tuning or additional data collection.

We also evaluate the contribution of the instruction refinement module independently. Using only the Qwen3.5-397B-A17B refinement model, our system achieves an average score of 69.7, which already exceeds the performance of the original Qwen3.5-397B-A17B model (65.6) on the same benchmark. When combined with the conditional zoom-in mechanism, the performance further increases to 76.8. 

Notably, this improvement is achieved by introducing only an additional 4B grounding model alongside the 397B refinement model. This result suggests that the proposed framework can effectively leverage the reasoning capability of large models while delegating precise localization to a smaller specialized grounding model. Consequently, our approach provides a cost-efficient way to enhance the GUI grounding capability of existing large models without requiring extensive fine-tuning or additional large-scale data collection.
\section{Conclusion}

\paragraph{}
In this work, we present AdaZoom-GUI, a novel framework for GUI grounding that addresses the challenges of high-resolution screenshots, small UI elements, and ambiguous user instructions. Our approach decomposes the task into two complementary components: an instruction refinement module that converts natural language tasks into explicit, visually grounded descriptions, and a grounding model that predicts both click coordinates and bounding boxes of target elements. To further improve localization accuracy while maintaining efficiency, we introduced a conditional zoom-in strategy that selectively performs a second-stage inference when the predicted element size indicates that a closer inspection is necessary.

To support this framework, we constructed a high-quality GUI grounding dataset and trained the grounding model using Group Relative Policy Optimization (GRPO), enabling the model to jointly optimize click-point accuracy and bounding box prediction. Extensive experiments on public benchmarks demonstrate that the proposed method significantly improves grounding performance compared with existing approaches. In particular, our method achieves competitive or superior results to models with comparable or even larger parameter sizes, while maintaining an efficient inference pipeline.

Overall, our results highlight the importance of separating instruction comprehension from grounding and adapting inference strategies based on element scale. These findings suggest a practical and scalable direction for improving GUI agents and enabling more reliable interaction with real-world graphical interfaces. Future work may explore extending this framework to multi-step GUI tasks and broader interactive agent settings.

{
    \small
    \bibliographystyle{ieeenat_fullname}
    \bibliography{main}

@article{wang2024gui,
  title={Gui agents with foundation models: A comprehensive survey},
  author={Wang, Shuai and Liu, Weiwen and Chen, Jingxuan and Zhou, Yuqi and Gan, Weinan and Zeng, Xingshan and Che, Yuhan and Yu, Shuai and Hao, Xinlong and Shao, Kun and others},
  journal={arXiv preprint arXiv:2411.04890},
  year={2024}
}

@article{kim2023language,
  title={Language models can solve computer tasks},
  author={Kim, Geunwoo and Baldi, Pierre and McAleer, Stephen},
  journal={Advances in Neural Information Processing Systems},
  volume={36},
  pages={39648--39677},
  year={2023}
}

@inproceedings{cheng2024seeclick,
  title={Seeclick: Harnessing gui grounding for advanced visual gui agents},
  author={Cheng, Kanzhi and Sun, Qiushi and Chu, Yougang and Xu, Fangzhi and YanTao, Li and Zhang, Jianbing and Wu, Zhiyong},
  booktitle={Proceedings of the 62nd Annual Meeting of the Association for Computational Linguistics (Volume 1: Long Papers)},
  pages={9313--9332},
  year={2024}
}

@article{tang2025gui,
  title={GUI-G$^2$: Gaussian Reward Modeling for GUI Grounding},
  author={Tang, Fei and Gu, Zhangxuan and Lu, Zhengxi and Liu, Xuyang and Shen, Shuheng and Meng, Changhua and Wang, Wen and Zhang, Wenqi and Shen, Yongliang and Lu, Weiming and others},
  journal={arXiv preprint arXiv:2507.15846},
  year={2025}
}

@article{yang2025gta1,
  title={Gta1: Gui test-time scaling agent},
  author={Yang, Yan and Li, Dongxu and Dai, Yutong and Yang, Yuhao and Luo, Ziyang and Zhao, Zirui and Hu, Zhiyuan and Huang, Junzhe and Saha, Amrita and Chen, Zeyuan and others},
  journal={arXiv preprint arXiv:2507.05791},
  year={2025}
}

@misc{gemini3pro_modelcard,
  title={Gemini 3 Pro - Model Card},
  author={{Google DeepMind}},
  year={2025},
  url={https://storage.googleapis.com/deepmind-media/Model-Cards/Gemini-3-Pro-Model-Card.pdf},
}

@article{Qwen3-VL,
      title={Qwen3-VL Technical Report}, 
      author={Shuai Bai and Yuxuan Cai and Ruizhe Chen and Keqin Chen and Xionghui Chen and Zesen Cheng and Lianghao Deng and Wei Ding and Chang Gao and Chunjiang Ge and Wenbin Ge and Zhifang Guo and Qidong Huang and Jie Huang and Fei Huang and Binyuan Hui and Shutong Jiang and Zhaohai Li and Mingsheng Li and Mei Li and Kaixin Li and Zicheng Lin and Junyang Lin and Xuejing Liu and Jiawei Liu and Chenglong Liu and Yang Liu and Dayiheng Liu and Shixuan Liu and Dunjie Lu and Ruilin Luo and Chenxu Lv and Rui Men and Lingchen Meng and Xuancheng Ren and Xingzhang Ren and Sibo Song and Yuchong Sun and Jun Tang and Jianhong Tu and Jianqiang Wan and Peng Wang and Pengfei Wang and Qiuyue Wang and Yuxuan Wang and Tianbao Xie and Yiheng Xu and Haiyang Xu and Jin Xu and Zhibo Yang and Mingkun Yang and Jianxin Yang and An Yang and Bowen Yu and Fei Zhang and Hang Zhang and Xi Zhang and Bo Zheng and Humen Zhong and Jingren Zhou and Fan Zhou and Jing Zhou and Yuanzhi Zhu and Ke Zhu},
	  journal={arXiv preprint arXiv:2511.21631},
      year={2025}
}

@inproceedings{park2025r,
  title={R-vlm: Region-aware vision language model for precise gui grounding},
  author={Park, Joonhyung and Tang, Peng and Das, Sagnik and Appalaraju, Srikar and Singh, Kunwar Yashraj and Manmatha, R and Ghadar, Shabnam},
  booktitle={Findings of the Association for Computational Linguistics: ACL 2025},
  pages={9669--9685},
  year={2025}
}

@article{zhou2025mai,
  title={MAI-UI Technical Report: Real-World Centric Foundation GUI Agents},
  author={Zhou, Hanzhang and Zhang, Xu and Tong, Panrong and Zhang, Jianan and Chen, Liangyu and Kong, Quyu and Cai, Chenglin and Liu, Chen and Wang, Yue and Zhou, Jingren and others},
  journal={arXiv preprint arXiv:2512.22047},
  year={2025}
}

@article{shao2024deepseekmath,
  title={Deepseekmath: Pushing the limits of mathematical reasoning in open language models},
  author={Shao, Zhihong and Wang, Peiyi and Zhu, Qihao and Xu, Runxin and Song, Junxiao and Bi, Xiao and Zhang, Haowei and Zhang, Mingchuan and Li, YK and Wu, Yang and others},
  journal={arXiv preprint arXiv:2402.03300},
  year={2024}
}

@inproceedings{kapoor2024omniact,
  title={Omniact: A dataset and benchmark for enabling multimodal generalist autonomous agents for desktop and web},
  author={Kapoor, Raghav and Butala, Yash Parag and Russak, Melisa and Koh, Jing Yu and Kamble, Kiran and AlShikh, Waseem and Salakhutdinov, Ruslan},
  booktitle={European Conference on Computer Vision},
  pages={161--178},
  year={2024},
  organization={Springer}
}

@article{gou2024navigating,
  title={Navigating the digital world as humans do: Universal visual grounding for gui agents},
  author={Gou, Boyu and Wang, Ruohan and Zheng, Boyuan and Xie, Yanan and Chang, Cheng and Shu, Yiheng and Sun, Huan and Su, Yu},
  journal={arXiv preprint arXiv:2410.05243},
  year={2024}
}

@misc{zhao2024swiftascalablelightweightinfrastructure,
      title={SWIFT:A Scalable lightWeight Infrastructure for Fine-Tuning},
      author={Yuze Zhao and Jintao Huang and Jinghan Hu and Xingjun Wang and Yunlin Mao and Daoze Zhang and Zeyinzi Jiang and Zhikai Wu and Baole Ai and Ang Wang and Wenmeng Zhou and Yingda Chen},
      year={2024},
      eprint={2408.05517},
      archivePrefix={arXiv},
      primaryClass={cs.CL},
      url={https://arxiv.org/abs/2408.05517},
}

@article{hurst2024gpt,
  title={Gpt-4o system card},
  author={Hurst, Aaron and Lerer, Adam and Goucher, Adam P and Perelman, Adam and Ramesh, Aditya and Clark, Aidan and Ostrow, AJ and Welihinda, Akila and Hayes, Alan and Radford, Alec and others},
  journal={arXiv preprint arXiv:2410.21276},
  year={2024}
}

@misc{Claude3_7_sonnet,
  title={Claude 3.7 sonnet},
  author={Anthropic},
  year={2025},
  url={https://www.anthropic.com/news/claude-3-7-sonnet},
}

@misc{Operater,
  title={Operater},
  author={OpenAI},
  year={2025},
  url={https://openai.com/index/introducing-operator/},
}

@techreport{seed2025seed1,
  title={Seed1.8 model card: Towards generalized real-world agency},
  author={Seed, Bytedance},
  year={2025},
}

@misc{qwen3.5,
    title  = {{Qwen3.5}: Towards Native Multimodal Agents},
    author = {{Qwen Team}},
    month  = {February},
    year   = {2026},
    url    = {https://qwen.ai/blog?id=qwen3.5}
}

@article{liu2025infigui,
  title={Infigui-r1: Advancing multimodal gui agents from reactive actors to deliberative reasoners},
  author={Liu, Yuhang and Li, Pengxiang and Xie, Congkai and Hu, Xavier and Han, Xiaotian and Zhang, Shengyu and Yang, Hongxia and Wu, Fei},
  journal={arXiv preprint arXiv:2504.14239},
  year={2025}
}

@article{wu2025gui,
  title={Gui-actor: Coordinate-free visual grounding for gui agents},
  author={Wu, Qianhui and Cheng, Kanzhi and Yang, Rui and Zhang, Chaoyun and Yang, Jianwei and Jiang, Huiqiang and Mu, Jian and Peng, Baolin and Qiao, Bo and Tan, Reuben and others},
  journal={arXiv preprint arXiv:2506.03143},
  year={2025}
}

@article{wang2025opencua,
  title={Opencua: Open foundations for computer-use agents},
  author={Wang, Xinyuan and Wang, Bowen and Lu, Dunjie and Yang, Junlin and Xie, Tianbao and Wang, Junli and Deng, Jiaqi and Guo, Xiaole and Xu, Yiheng and Wu, Chen Henry and others},
  journal={arXiv preprint arXiv:2508.09123},
  year={2025}
}

@article{ye2025mobile,
  title={Mobile-agent-v3: Fundamental agents for gui automation},
  author={Ye, Jiabo and Zhang, Xi and Xu, Haiyang and Liu, Haowei and Wang, Junyang and Zhu, Zhaoqing and Zheng, Ziwei and Gao, Feiyu and Cao, Junjie and Lu, Zhengxi and others},
  journal={arXiv preprint arXiv:2508.15144},
  year={2025}
}

@misc{Holo2,
  title={Holo2: Foundational Models for Navigation and Computer Use Agents},
  author={H Company},
  year={2025},
  url={https://huggingface.co/Hcompany/Holo2-4B},
}

@article{wu2024atlas,
  title={Os-atlas: A foundation action model for generalist gui agents},
  author={Wu, Zhiyong and Wu, Zhenyu and Xu, Fangzhi and Wang, Yian and Sun, Qiushi and Jia, Chengyou and Cheng, Kanzhi and Ding, Zichen and Chen, Liheng and Liang, Paul Pu and others},
  journal={arXiv preprint arXiv:2410.23218},
  year={2024}
}

@inproceedings{
    li2025screenspotpro,
    title={ScreenSpot-Pro: {GUI} Grounding for Professional High-Resolution Computer Use},
    author={Kaixin Li and Meng Ziyang and Hongzhan Lin and Ziyang Luo and Yuchen Tian and Jing Ma and Zhiyong Huang and Tat-Seng Chua},
    booktitle={Workshop on Reasoning and Planning for Large Language Models},
    year={2025},
    url={https://openreview.net/forum?id=XaKNDIAHas}
}
}


\end{document}